\documentclass[letterpaper, 10 pt, conference]{ieeeconf}  % Comment this line out if you need a4paper

\IEEEoverridecommandlockouts                              % This command is only needed if 
\usepackage{graphics} % for pdf, bitmapped graphics files
\usepackage{epsfig} % for postscript graphics files
\usepackage{amsmath} % assumes amsmath package installed
\usepackage{amssymb}  % assumes amsmath package installed
\usepackage{mathtools}
\usepackage{algorithm}
\usepackage{algpseudocode} % algorithmicx
\usepackage{multirow}
\usepackage{booktabs}
\usepackage[caption=false,font=footnotesize]{subfig}

\title{\LARGE \bf
Clustering-Embedded Model Predictive Path Integral Control: Avoiding Averaging-Induced Failure and Enabling Efficient Cluster Selection for Dynamic Obstacles
}

\author{Zidong Liu$^{1}$, Kaixin Chang$^{1}$ and Xu Chen$^{\dagger, 1}$% <-this % stops a space
\thanks{*This work was supported in part by an Amazon-UW Science Hub Gift Award. The views and conclusions
contained in this document are those of the authors and should not be interpreted as representing the official policies,
either expressed or implied, of the sponsoring organization.
}% <-this % stops a space
\thanks{$^\dagger$: corresponding author.}
\thanks{$^{1}$Department of Mechanical Engineering, University of Washington,  Seattle, WA, USA. {\tt\small\{liuzd, kxc118, chx\}@uw.edu}}%
}

\begin{document}

\maketitle
\thispagestyle{empty}
\pagestyle{empty}

%%%%%%%%%%%%%%%%%%%%%%%%%%%%%%%%%%%%%%%%%%%%%%%%%%%%%%%%%%%%%%%%%%%%%%%%%%%%%%%%
\begin{abstract}
With the widespread availability of parallel computing hardware, sampling-based motion planning methods such as Model Predictive Path Integral (MPPI) control have become increasingly powerful for complex nonlinear systems in non-smooth task spaces. However, the sampling and forward-simulation pipeline in MPPI suffers from averaging-induced failure in cluttered environments, where the importance-weighted update averages incompatible rollouts and leads to hesitation or even collision when an obstacle lies directly ahead. 
This paper proposes Clustering-Embedded MPPI (CE-MPPI), a framework that architecturally resolves the averaging-induced failures inherent in standard MPPI within non-convex environments. Rather than simply mitigating interference, CE-MPPI redefines the control law by integrating a high-fidelity pruning and clustering stage.
By leveraging density-based spatial clustering of applications with noise (DBSCAN) alongside a novel geometric direction feature that is extracted from collision-derived reference points, the system isolates feasible trajectory modes from the noise of infeasible rollouts. This is paired with an intelligent selection logic that optimizes for minimum cost in static scenes while actively steering opposite to obstacle flux in dynamic environments.
% This paper proposes Clustering-Enhanced MPPI (CE-MPPI) to mitigate this issue. CE-MPPI first prunes colliding rollouts and then clusters the remaining feasible rollouts using density-based spatial clustering of applications with noise (DBSCAN) with a geometric direction feature defined by the terminal displacement from a collision-derived reference point computed from colliding rollouts near the closest obstacle. For static obstacles, CE-MPPI selects the cluster with the minimum average cost; for dynamic obstacles, it selects the cluster whose direction is most opposite to the estimated obstacle motion direction from recent observations. The nominal control sequence is then updated using a within-cluster path-integral weighted update. 
%Experiments in 2-D simulations accelerated with JAX and real-world tests on a 6-DoF UR5e manipulator with CUDA-parallel rollouts in Isaac Gym demonstrate that CE-MPPI effectively mitigates averaging-induced failure and reduces persistent interference from obstacles moving in similar directions to the robot, improving planning efficiency and overall performance.
Experiments in 2-D JAX-accelerated simulations show that CE-MPPI alleviates obstacle-front hesitation and avoids persistent coupling with moving obstacles in dynamic scenes. In particular, real-world tests on a 6-DoF UR5e manipulator with CUDA-parallel rollouts in Isaac Gym achieve a 48\% reduction in time-to-goal and a 12\% shorter end-effector path.

\end{abstract}

%%%%%%%%%%%%%%%%%%%%%%%%%%%%%%%%%%%%%%%%%%%%%%%%%%%%%%%%%%%%%%%%%%%%%%%%%%%%%%%%
\section{INTRODUCTION}
With the rapid growth of computing hardware, sampling-based motion planning methods~\cite{kazim2024recent,orthey2023sampling} such as Model Predictive Path Integral (MPPI) control~\cite{williams2018information} have become increasingly popular for real-time robotic motion planning. MPPI iteratively generates stochastic rollouts, performs forward simulation, evaluates trajectory costs, and applies a path-integral weighted update to refine the control sequence. Compared to traditional gradient-based MPC~\cite{schwenzer2021review}, MPPI does not require the cost function to be differentiable, making it convenient to deploy with nonlinear dynamics and complex, non-smooth objectives~\cite{bhardwaj2022storm}. Moreover, unlike learning-based approaches such as reinforcement learning~\cite{tang2025deep} that typically rely on extensive offline training, sampling-based motion planning operates as a fully online planner and can further benefit from GPU-accelerated parallel rollout computation~\cite{lee2024gpu} for real-time performance.

The performance of sampling-based motion planning hinges on the sampling number and the diversity of rollouts~\cite{razmjoo2024sampling}. For example for MPPI, at each replanning step, the algorithm samples control sequences by injecting stochastic perturbations (typically Gaussian noise) around a nominal control sequence and then applies an importance-weighted update. However, the weighted averaging mechanism can be problematic in cluttered environments with multi-modal feasible solutions~\cite{park2025csc}. As illustrated in Fig.~1, even when colliding rollouts receive negligible weights, importance-weighted averaging of left- and right-passing rollouts can produce a control update toward the obstacle, causing hesitation and potential collision, leading to what we shall refer to as averaging-induced failure.

\begin{figure}[htbp]
    \centering
    \includegraphics[width=0.6\linewidth]{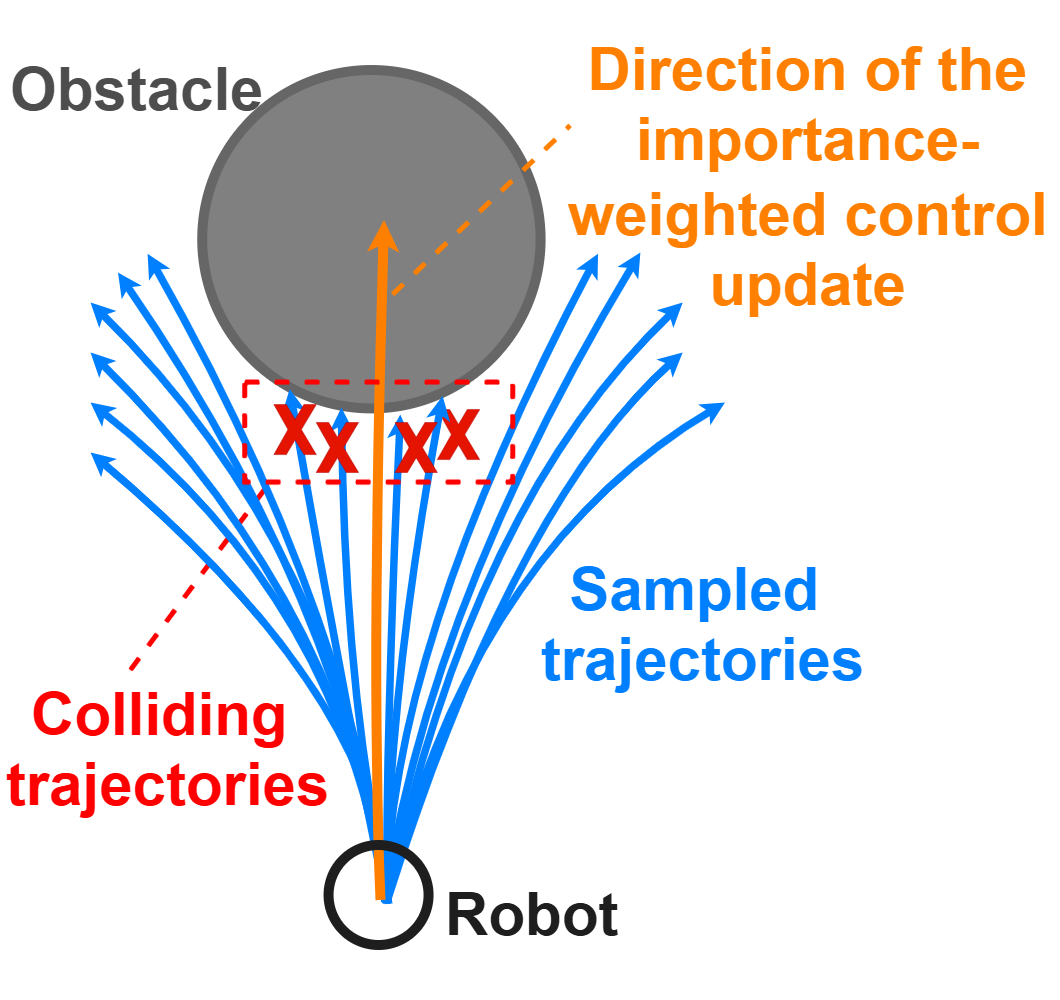}
    \caption{Averaging-induced failure in MPPI: the weighted control update may point toward the obstacle even when colliding rollouts are heavily penalized.}
    \label{CollidingTraj}
\end{figure}

To address averaging-induced failure and maximize the sampling efficiency in motion planning, we propose Clustering-Embedded MPPI (CE-MPPI), as illustrated in Fig.~\ref{Framework}. Exploiting the massive parallelism of GPU-accelerated forward simulations, CE-MPPI engineers Gaussian control perturbations into a high-speed optimization engine.
After rollouts are generated, we first prune colliding trajectories. For the remaining feasible rollouts, we apply the density-based spatial clustering of application with noise (DBSCAN) clustering~\cite{deng2020dbscan} using a compact geometric feature defined by the direction from a collision-derived reference point to each rollout terminal state, where the reference point is computed from the mean terminal position of colliding rollouts near the closest obstacle. This feature yields well-separated clusters corresponding to distinct avoidance directions (e.g., left vs.\ right in 2-D). CE-MPPI then performs obstacle-aware cluster selection. For static obstacles, we select the cluster with the minimum average cost, which avoids averaging across incompatible modes while prioritizing low-cost progress, as illustrated in Fig.~\ref{StaticObs}. For dynamic obstacles, we estimate the obstacle motion direction from recent observations and select the cluster whose average motion is most opposite to the obstacle motion, based on a dot-product criterion, as shown in Fig.~\ref{DynamicObs}. This directional selection mitigates persistent interference when the robot and obstacle move in similar directions, where greedy cost reduction can keep the robot coupled with the moving obstacle for an extended period. Finally, CE-MPPI updates the nominal control sequence using only rollouts within the selected cluster.

We validate CE-MPPI in JAX~\cite{bradbury2018jax} with Just-In-Time (JIT) acceleration in 2-D simulations, demonstrating 
% effective avoidance of averaging-induced failure and improved replanning in dynamic scenes through early bypassing guided by obstacle motion direction. 
a systematic resolution of averaging-induced failures and achieving robust replanning in dynamic scenes through proactive bypassing guided by obstacle motion direction.
We further deploy CE-MPPI on a real 6-(Degrees of Freedom) DoF UR5e manipulator with CUDA-accelerated rollouts in Isaac Gym~\cite{makoviychuk2021isaac}, showing that CE-MPPI remains efficient and effective for high-DoF obstacle avoidance in real-world settings.

\subsection{Related Work}
MPPI was initially demonstrated in challenging, high-speed autonomous driving tasks~\cite{williams2018information}. Since then, it has been adopted in a broad range of robotic systems, including manipulators~\cite{bhardwaj2022storm,vasilopoulos2023ramp,zhou2025parallel}, drones~\cite{zhai2026pa}, and legged robots~\cite{alvarez2025real}.
STORM~\cite{bhardwaj2022storm} and RAMP~\cite{vasilopoulos2023ramp} demonstrated MPPI-style planning on 6--7 DoF manipulators with GPU acceleration, enabling real-time goal reaching and obstacle avoidance in cluttered workspaces.
Zhou \emph{et al.}~\cite{zhou2025parallel} further improved real-time performance for dynamic obstacle avoidance of manipulators by proposing a parallel MPPI framework with a gradient-velocity modulated signed distance field (SDF) cost. Zhai \emph{et al.}~\cite{zhai2026pa} proposed Perception-Aware MPPI for quadrotor navigation in unknown environments by augmenting MPPI with perception-driven objectives to encourage online exploration and alternative path discovery when the goal is occluded. Alvarez-Padilla \emph{et al.}~\cite{alvarez2025real} demonstrated real-time whole-body control on a real-world legged robot using MPPI.

% A number of MPPI variants have been proposed to improve robustness and practical performance, including robust MPPI that augments the state space~\cite{gandhi2021robust}, smooth MPPI that adds action regularization and input-lifting to suppress control chattering~\cite{kim2022smooth}, and U-MPPI that leverages unscented guidance to propagate uncertainty and improve sampling efficiency with risk-aware trajectory evaluation~\cite{mohamed2025toward}

A major computational bottleneck of MPPI lies in sampling and the associated parallel forward simulations, motivating extensive research on improving sampling efficiency and the rollout generation process. Kim \emph{et al.}~\cite{kim2025single} proposed a single-instance sampling strategy for task-space MPPI control to reduce computational burden while maintaining real-time accuracy. Yan and Devasia~\cite{yan2024output} introduced output-sampled MPPI (o-MPPI), which samples in an output space to better satisfy output constraints and achieve higher efficiency with fewer rollouts and shorter horizons. Aoki \emph{et al.}~\cite{aoki2024switching} further studied switching the sampling space online to balance efficiency and safety in vehicle navigation. In the context of mitigating averaging-induced failure, CSC-MPPI~\cite{park2025csc} addressed weighted-averaging issues by first enforcing constraints using a primal--dual gradient-based adjustment that iteratively shifts sampled trajectories into feasible regions, and then clustering the resulting feasible rollouts via DBSCAN using features derived from average linear and angular velocities. There, the cluster with the minimum average cost was selected for the path-integral update. However, because velocity-based statistics can be similar across distinct avoidance directions, left- and right-passing clusters may remain weakly separated, which can degrade clustering quality and lead to hesitation near obstacles. Moreover, in dynamic scenes where obstacle motion is aligned with the robot's motion, selecting clusters purely based on instantaneous cost may still result in persistent coupling with the moving obstacle. %/计算时间

\subsection{Contributions}
The main contributions of this paper are as follows:
    \begin{enumerate}
        %\item CE-MPPI is proposed to prune colliding rollouts and cluster feasible rollouts via DBSCAN using a geometric direction feature defined by the terminal displacement from a collision-derived reference point, improving the accuracy and reliability of rollout clustering.
        %\item An obstacle-aware cluster selection strategy is introduced: for static obstacles, CE-MPPI selects the cluster with the minimum average cost; for dynamic obstacles, it selects the cluster whose average motion is opposite to the obstacle motion direction. This mitigates averaging-induced failure and reduces persistent blocking and collision risk in dynamic scenes.
        % \item We propose CE-MPPI, which engineers a high-fidelity pruning-and-clustering stage that removes colliding rollouts and clusters feasible rollouts via DBSCAN. By introducing a novel geometric direction feature anchored to collision-derived reference points, the framework guarantees the isolation of distinct, feasible trajectory modes, ensuring the integrity of the optimization manifold.
        \item We propose CE-MPPI, which engineers a high-fidelity pruning-and-clustering stage that removes colliding rollouts and clusters feasible rollouts via DBSCAN. By introducing a novel geometric direction feature anchored to collision-derived reference points, the framework separates distinct feasible trajectory modes and enables a mode-consistent update.
        \item We provide a decisive obstacle-aware selection logic: while optimizing for minimum cost in static environments, the algorithm executes an anticipatory response in dynamic scenes by selecting clusters that counter obstacle flux. This approach resolves averaging-induced failures and preempts the persistent blocking risks inherent in unpredictable physical spaces.
        % \item CE-MPPI is validated through comparative experiments against MPPI and CSC-MPPI in both 2-D simulations with JAX acceleration and real-world tests on a 6-DoF UR5e manipulator using CUDA-accelerated rollouts in Isaac Gym, demonstrating a step change in efficiency and performance for real-time motion planning.
        \item CE-MPPI is validated through comparative experiments against MPPI and CSC-MPPI in both simulation and real-world experiments. In 2-D JAX-accelerated simulations, CE-MPPI mitigates averaging-induced failure and demonstrates strong escape capability from persistent coupling with moving obstacles, reducing time-to-goal by 17.4\% and path length by 9.1\% compared to CSC-MPPI in the dynamic scenario. In real-world experiments on a 6-DoF UR5e manipulator with CUDA-accelerated rollouts in Isaac Gym, CE-MPPI achieves a 48\% reduction in time-to-goal and a 12\% shorter end-effector path compared to standard MPPI.
    \end{enumerate}

\section{Proposed Clustering-Embedded MPPI}

\begin{figure*}[t]
    \centering
    \includegraphics[width=0.9\textwidth]{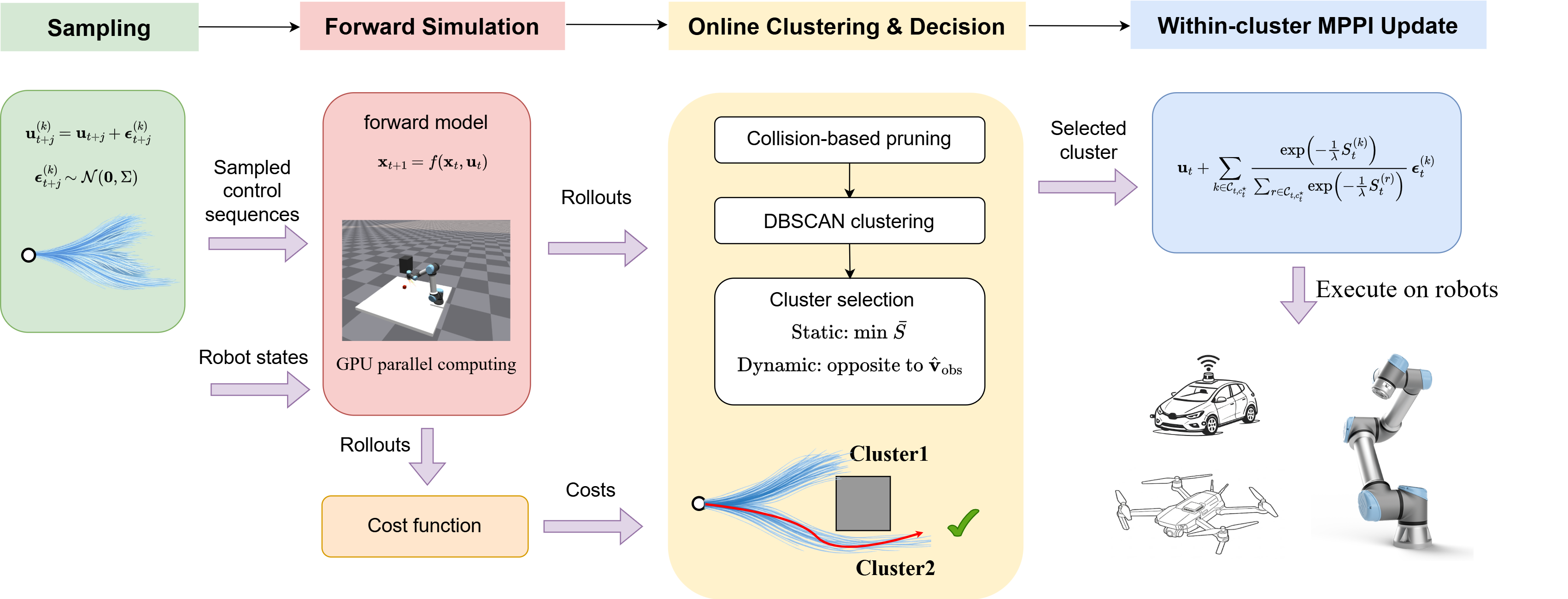}
    \caption{Overview of the proposed CE-MPPI. CE-MPPI samples perturbed control sequences and performs parallel forward simulation to obtain trajectory rollouts and costs. Colliding rollouts are pruned, the remaining feasible rollouts are clustered via DBSCAN, and a single cluster is selected (static: minimum average cost; dynamic: motion opposite to the obstacle direction). The MPPI update is then computed using weights normalized over the selected cluster.}
    \label{Framework}
\end{figure*}

\subsection{MPPI Preliminaries}
\label{subsec:mppi-preliminaries}
We consider a general discrete-time system
\begin{equation}
\mathbf{x}_{t+1} = f(\mathbf{x}_t, \mathbf{u}_t) ,
\end{equation}
where $\mathbf{x}_t \in \mathbb{R}^{n}$ and $\mathbf{u}_t \in \mathbb{R}^{m}$ denote the system state and control input, respectively. At each iteration $t$, MPPI optimizes a horizon-$N$ control sequence
\begin{equation}
\mathbf{U}_t \coloneqq \{\mathbf{u}_{t+j}\}_{j=0}^{N-1},
\end{equation}
where $j\in\{0,\dots,N-1\}$ indexes the within-horizon time step.

MPPI evaluates $K$ stochastic rollouts by perturbing the nominal controls.
More specifically, for rollout index $k\in\{1,\dots,K\}$, sampled Gaussian noise
$\boldsymbol{\epsilon}^{(k)}_{t+j}\sim\mathcal{N}(\mathbf{0},\Sigma)$
is added to form the perturbed control
\begin{equation}
\mathbf{u}^{(k)}_{t+j} = \mathbf{u}_{t+j} + \boldsymbol{\epsilon}^{(k)}_{t+j}, \quad j=0,\dots,N-1.
\end{equation}
Starting from $\mathbf{x}^{(k)}_{t}=\mathbf{x}_t$, the rollout is obtained via forward simulation
\begin{equation}
\mathbf{x}^{(k)}_{t+j+1}=f\!\left(\mathbf{x}^{(k)}_{t+j},\mathbf{u}^{(k)}_{t+j}\right),\quad j=0,\dots,N-1.
\label{eq:fk}
\end{equation}

Each rollout is assigned a trajectory cost
\begin{equation}
S^{(k)}_t = \phi\!\left(\mathbf{x}^{(k)}_{t+N}\right) + \sum_{j=0}^{N-1} \ell\!\left(\mathbf{x}^{(k)}_{t+j},\mathbf{u}^{(k)}_{t+j}\right),
\label{eq:cost}
\end{equation}
where $\phi(\cdot)$ and $\ell(\cdot)$ denote terminal and stage costs, respectively.
The importance weight for rollout $k$ is then defined as
\begin{equation}
w^{(k)}_t=
\frac{\exp\!\left(-\frac{1}{\lambda}S^{(k)}_t\right)}
{\sum_{r=1}^{K}\exp\!\left(-\frac{1}{\lambda}S^{(r)}_t\right)},
\label{eq:mppi-w}
\end{equation}
where $\lambda>0$ is the temperature parameter controlling the sharpness of the weighting.

Finally, MPPI updates the nominal controls via a weighted average of perturbations:
\begin{equation}
\mathbf{u}_{t+j} \coloneqq \mathbf{u}_{t+j} + \sum_{k=1}^{K} w^{(k)}_t\,\boldsymbol{\epsilon}^{(k)}_{t+j},\quad j=0,\dots,N-1.
\label{eq:control_update}
\end{equation}
Only the first control $\mathbf{u}_{t}$ is executed, and the procedure repeats in a receding-horizon manner. 

One can note that in cluttered environments with dynamic moving objects, the standard MPPI is subjected to averaging-induced failure. Although multiple feasible modes can be sampled, the importance-weighted averaging step can blend incompatible rollouts, producing a control update that points toward collision. 

\subsection{Proposed CE-MPPI}
\label{subsec:ce-mppi}
% The proposed CE-MPPI that filters out colliding rollouts and clusters the remaining feasible rollouts via DBSCAN. The controller then selects one cluster, choosing the lowest-cost cluster for static obstacles. For dynamic obstacles, CE-MPPI additionally selects the cluster whose average motion opposes the obstacle motion direction, which helps avoid persistent blocking when the robot moves in a similar direction to the obstacle over the horizon. Finally, the nominal control sequence is updated using only rollouts from the selected cluster.

Fig.~\ref{Framework} illustrates the overall workflow of the proposed CE-MPPI, and Alg.~\ref{alg:cemppi} provides the corresponding pseudocode. 
% CE-MPPI first filters out colliding rollouts and clusters the remaining feasible rollouts via DBSCAN. The controller then selects one cluster, choosing the lowest-cost cluster for static obstacles. For dynamic obstacles, CE-MPPI additionally selects the cluster whose average motion opposes the obstacle motion direction, which helps avoid persistent blocking when the robot moves in a similar direction to the obstacle over the horizon. Finally, the nominal control sequence is updated using only rollouts from the selected cluster.
CE-MPPI redefines the control loop by first pruning colliding trajectories and distilling the remaining feasible rollouts into distinct motion modes via DBSCAN. The system then executes a decisive selection logic: 1) static environments trigger an optimization for the absolute minimum cost, and 2) dynamic scenes evoke an anticipatory strategy that selects clusters countering obstacle flux. By isolating these high-integrity rollouts, the nominal control sequence is updated to eliminate the risk of persistent blocking and averaging-induced failure.
Next, we detail each main stage of the process: collision-based pruning, DBSCAN clustering, obstacle-aware cluster selection, and the within-cluster MPPI update.

\begin{figure}[htbp]
    \centering
    \includegraphics[width=0.7\linewidth]{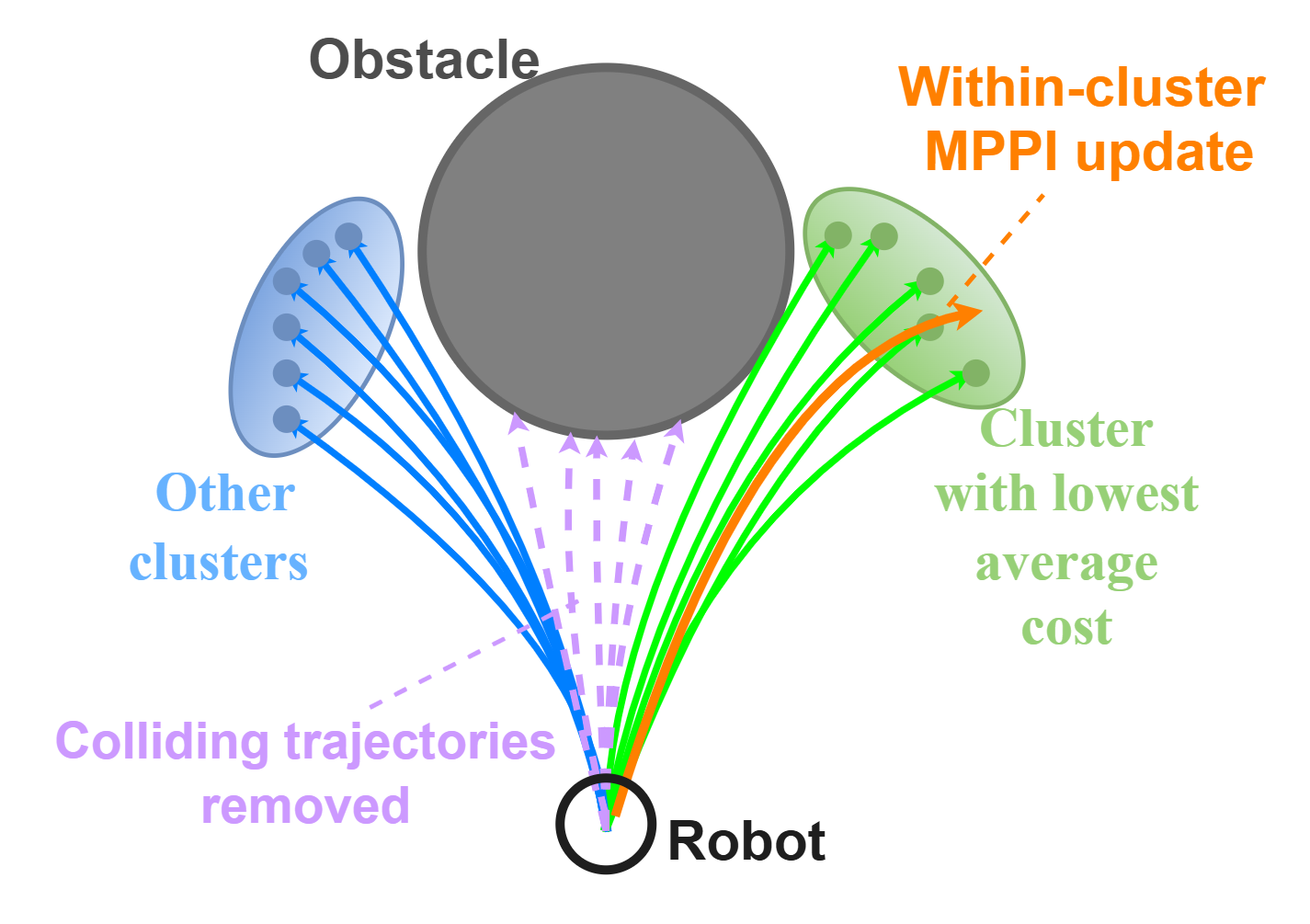}
    \caption{Static obstacle case in CE-MPPI. Colliding rollouts are pruned (purple dashed trajectories), and the remaining feasible rollouts are clustered into distinct avoidance clusters. CE-MPPI selects the cluster with the minimum average cost and performs the within-cluster MPPI update to compute the control input.}
    \label{StaticObs}
\end{figure}

\begin{algorithm}[t]
\caption{Proposed CE-MPPI}
\label{alg:cemppi}
\begin{algorithmic}[1]
\Require $\mathbf{x}_t$, $\mathbf{U}_t=\{\mathbf{u}_{t+j}\}_{j=0}^{N-1}$, $K,N,\lambda,\Sigma$, $f(\cdot)$, $\ell(\cdot),\phi(\cdot)$, $\mathbf{p}(\cdot)$, collision checker $\mathbb{I}_{\mathrm{col}}(\cdot)$, DBSCAN, $H,\Delta t$, $\{\mathbf{o}_{t-h}\}_{h=0}^{H-1}$ (if dynamic)
\Ensure updated $\mathbf{U}_t$ (apply $\mathbf{u}_t$)

\State \textbf{[Rollouts]} \textbf{for each} $k=1,\dots,K$ \textbf{in parallel do}
\State \hspace{0.8em} sample $\{\boldsymbol{\epsilon}^{(k)}_{t+j}\}_{j=0}^{N-1}$; set $\mathbf{u}^{(k)}_{t+j}\gets \mathbf{u}_{t+j}+\boldsymbol{\epsilon}^{(k)}_{t+j}$
\State \hspace{0.8em} simulate $\tau_t^{(k)}=\{\mathbf{x}^{(k)}_{t+j}\}_{j=0}^{N}$ via~\eqref{eq:fk}, compute $S_t^{(k)}$, and record $\mathbf{x}^{(k)}_{t+N}$
\State \hspace{0.8em} $c^{(k)}\gets \mathbb{I}_{\mathrm{col}}(\tau_t^{(k)})$
\State \textbf{end for}

\State \textbf{[Pruning]} $\mathcal{V}_t\gets\{k\mid c^{(k)}=0\}$,\quad $\mathcal{C}^{\mathrm{col}}_t\gets\{k\mid c^{(k)}=1\}$

\If{$|\mathcal{C}^{\mathrm{col}}_t|=0$}
  \State \textbf{[Fallback]} compute standard weights $w_t^{(k)}$ using~\eqref{eq:mppi-w}
  \State \textbf{[Update]} $\mathbf{u}_{t+j}\gets \mathbf{u}_{t+j}+\sum_{k=1}^{K} w_t^{(k)}\boldsymbol{\epsilon}^{(k)}_{t+j}$,\; $j=0,\dots,N-1$
\Else
  \State \textbf{[Clustering]} compute $\bar{\mathbf{p}}^{\,\mathrm{col}}_t$ and features $\mathbf{z}_t^{(k)}$ for $k\in\mathcal{V}_t$; run DBSCAN $\Rightarrow \{\mathcal{C}_{t,c}\}_{c=1}^{M_t}$ (ignore noise)
  \State \textbf{[Cluster stats]} $\bar{S}_{t,c}\gets \frac{1}{|\mathcal{C}_{t,c}|}\sum_{k\in\mathcal{C}_{t,c}} S_t^{(k)}$ for $c=1,\dots,M_t$

  \State \textbf{[Selection]} \textbf{if} static \textbf{then}
  \State \hspace{0.8em} $c_t^\star\gets \arg\min_c \bar{S}_{t,c}$
  \State \textbf{else} \Comment{dynamic obstacles}
  \State \hspace{0.8em} compute $\hat{\mathbf{v}}_{t,c}$ and $\hat{\mathbf{v}}_{\mathrm{obs},t}$
  \State \hspace{0.8em} $c_t^\star\gets \arg\min_c \hat{\mathbf{v}}_{t,c}^\top \hat{\mathbf{v}}_{\mathrm{obs},t}$
  \State \textbf{end if}

  \State \textbf{[Within-cluster weighting]} compute $\tilde{w}^{(k)}_t$ normalized over $\mathcal{C}_{t,c_t^\star}$ (e.g., using $\exp(-S/\lambda)$ and renormalization)
  \State \textbf{[Update]} $\mathbf{u}_{t+j}\gets \mathbf{u}_{t+j}+\sum_{k\in\mathcal{C}_{t,c_t^\star}}\tilde{w}^{(k)}_t\boldsymbol{\epsilon}^{(k)}_{t+j}$,\; $j=0,\dots,N-1$
\EndIf

\State \textbf{[Execute]} apply $\mathbf{u}_t$; shift and warm-start $\mathbf{U}_{t+1}$
\end{algorithmic}
\end{algorithm}

\subsubsection{Collision-based pruning of rollouts}
For each rollout $k\in\{1,\dots,K\}$ obtained by forward simulation in~\eqref{eq:fk}, we evaluate a collision indicator $\mathbb{I}_{\mathrm{col}}(\tau_t^{(k)})$,
where
\begin{equation}
    \tau_t^{(k)} := \{\mathbf{x}^{(k)}_{t+j}\}_{j=0}^{N}
\end{equation}
denotes the simulated state trajectory at replanning time $t$.
Rollouts that collide with obstacles are removed:
\begin{equation}
\mathcal{V}_t \coloneqq \left\{ k \,\middle|\, \mathbb{I}_{\mathrm{col}}\!\left(\tau_t^{(k)}\right)=0 \right\}\,,
\end{equation}
as illustrated by the purple dashed trajectories in Fig.~\ref{StaticObs}.
This pruning operation corresponds to the pruning stage in Alg.~\ref{alg:cemppi}. This step prevents invalid trajectories from receiving non-negligible weights and reduces the number of samples used in subsequent clustering and selection.

\subsubsection{DBSCAN clustering of valid rollouts}
This clustering procedure corresponds to the clustering stage in Alg.~\ref{alg:cemppi}. Rather than clustering full state sequences, we cluster feasible rollouts using a compact geometric feature derived from their terminal states $\mathbf{x}^{(k)}_{t+N}$.
Let $\mathbf{p}(\mathbf{x})\in\mathbb{R}^{p}$ denote a mapping of the state's task-relevant position (e.g., planar position for a mobile robot or end-effector position for a manipulator). We use the terminal positions of colliding rollouts to construct a collision-derived reference point.
Define the index set of colliding rollouts as
\begin{equation}
\mathcal{C}^{\mathrm{col}}_t \coloneqq \left\{ k \in \{1,\dots,K\} \,\middle|\, \mathbb{I}_{\mathrm{col}}(\tau_t^{(k)})=1 \right\}.
\end{equation}
When $|\mathcal{C}^{\mathrm{col}}_t|>0$, we compute the mean terminal position of colliding rollouts
\begin{equation}
\bar{\mathbf{p}}^{\,\mathrm{col}}_t \coloneqq \frac{1}{|\mathcal{C}^{\mathrm{col}}_t|}
\sum_{k\in\mathcal{C}^{\mathrm{col}}_t} \mathbf{p}\!\left(\mathbf{x}^{(k)}_{t+N}\right),
\end{equation}
which indicates where colliding rollouts tend to terminate.
For each feasible rollout $k\in\mathcal{V}_t$, we form a direction vector from this reference point to the rollout terminal position
\begin{equation}
\mathbf{d}^{(k)}_t \coloneqq \mathbf{p}\!\left(\mathbf{x}^{(k)}_{t+N}\right)-\bar{\mathbf{p}}^{\,\mathrm{col}}_t,
\end{equation}
and use its normalized direction as the clustering feature:
\begin{equation}
\mathbf{z}^{(k)}_t \coloneqq \frac{\mathbf{d}^{(k)}_t}{\|\mathbf{d}^{(k)}_t\|+\epsilon}.
\end{equation}
% We then apply DBSCAN to $\{\mathbf{z}^{(k)}_t\}_{k\in\mathcal{V}_t}$, which partitions the feasible rollouts into $M_t$ clusters $\{\mathcal{C}_{t,1},\dots,\mathcal{C}_{t,M_t}\}$, where $M_t$ is the number of clusters returned by DBSCAN. Noise rollouts identified by DBSCAN are ignored in subsequent cluster selection.

% Let $[\cdot]_x$ and $[\cdot]_y$ denote the coordinates of a vector projected onto a chosen 2-D task plane. We then convert $\mathbf{d}^{(k)}_t$ to an angle feature
% \begin{equation}
% \theta^{(k)}_t \coloneqq \operatorname{atan2}\!\left([\mathbf{d}^{(k)}_t]_y,\, [\mathbf{d}^{(k)}_t]_x\right).
% \end{equation}
% To avoid the discontinuity at $\pm\pi$, we embed the angle on the unit circle as
% \begin{equation}
% \mathbf{z}^{(k)}_t \coloneqq 
% \begin{bmatrix}
% \cos\theta^{(k)}_t\\
% \sin\theta^{(k)}_t
% \end{bmatrix},
% \end{equation}
% We then apply DBSCAN to $\{\mathbf{z}^{(k)}_t\}_{k\in\mathcal{V}_t}$, which partitions the feasible rollouts into $M_t$ clusters $\{\mathcal{C}_{t,1},\dots,\mathcal{C}_{t,M_t}\}$ with respect to the collision-derived reference point $\bar{\mathbf{p}}^{\,\mathrm{col}}_t$, where $M_t$ is the number of clusters returned by DBSCAN. Noise rollouts identified by DBSCAN are ignored in subsequent cluster selection.
We then apply DBSCAN to $\{\mathbf{z}^{(k)}_t\}_{k\in\mathcal{V}_t}$, yielding $M_t$ clusters $\{\mathcal{C}_{t,1},\dots,\mathcal{C}_{t,M_t}\}$, where $M_t$ is the number of clusters returned by DBSCAN. Noise rollouts identified by DBSCAN are ignored in subsequent cluster selection.
These clusters typically correspond to distinct local avoidance modes (e.g., passing an obstacle from different sides). The geometric feature is designed to promote cluster separability, enabling a mode-consistent MPPI update that avoids averaging across incompatible rollouts.

If no colliding rollout is observed (i.e., $|\mathcal{C}^{\mathrm{col}}_t|=0$), we skip clustering and revert to the standard MPPI update using all rollouts.

\subsubsection{Static and dynamic obstacle-aware cluster selection}
CE-MPPI selects a single cluster so that the MPPI update is computed from a coherent set of rollouts. This static criterion corresponds to the selection stage in Alg.~\ref{alg:cemppi}.

For static obstacles, we select the cluster with the lowest average cost:
\begin{equation}
c_t^{\star} \coloneqq \arg\min_{c\in\{1,\dots,M_t\}} \bar{S}_{t,c}\,,
\end{equation}
as illustrated by the green trajectories in Fig.~\ref{StaticObs}.

\begin{figure}[htbp]
    \centering
    \includegraphics[width=0.7\linewidth]{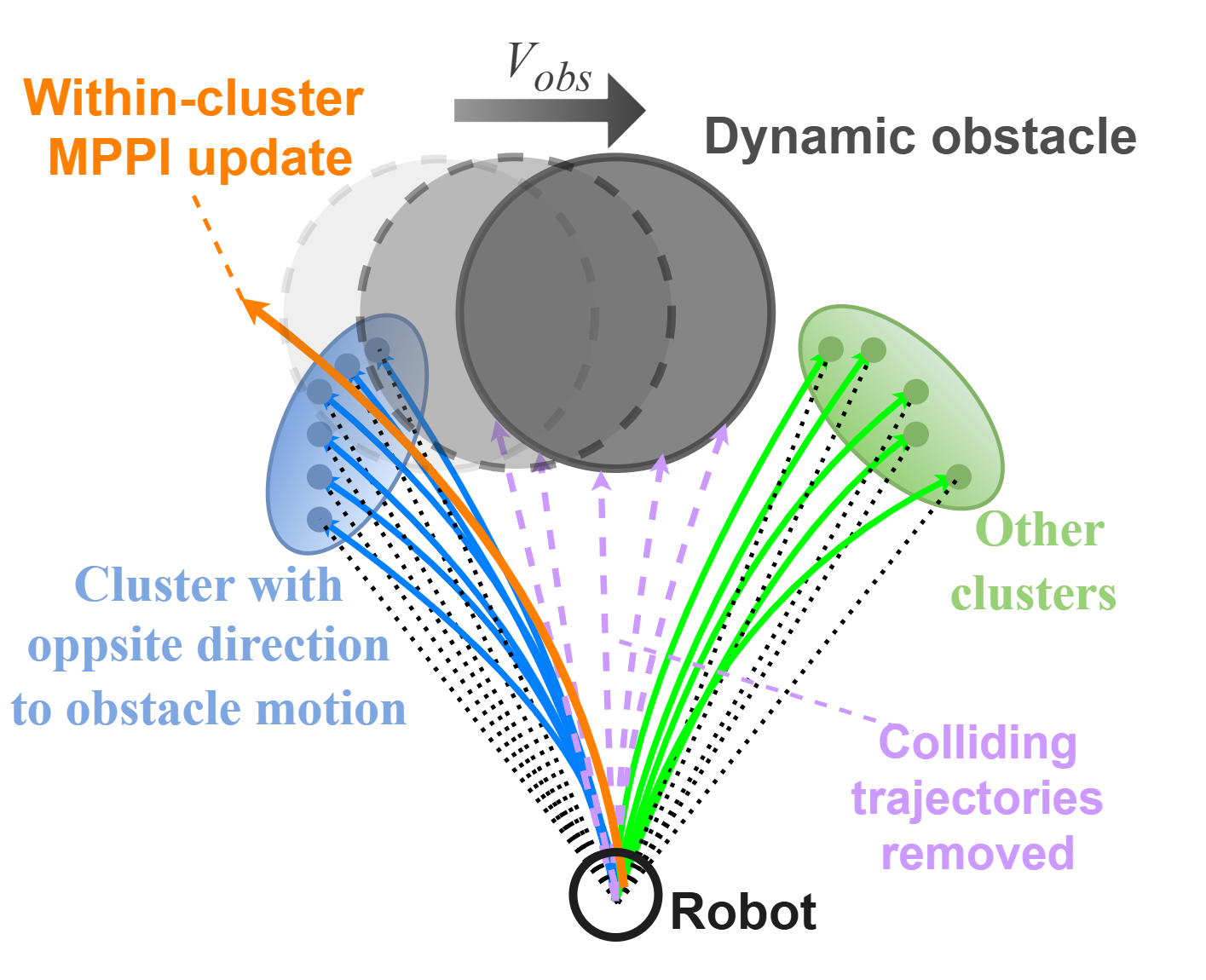}
    \caption{Dynamic obstacle case in CE-MPPI. Colliding rollouts are pruned (purple dashed trajectories), and the remaining feasible rollouts are clustered. Using the estimated obstacle motion direction $\mathbf{v}_{\mathrm{obs}}$, CE-MPPI selects the cluster whose average motion is most opposite to the obstacle motion and performs the within-cluster MPPI update to compute the control input.}
    \label{DynamicObs}
\end{figure}

For dynamic obstacles, shown as the gray circle in Fig.~\ref{DynamicObs}, we incorporate the estimated obstacle motion direction to guide the selection of the cluster. We first define the average motion direction of each cluster.
For rollout $k\in\mathcal{C}_{t,c}$, let
\begin{equation}
\mathbf{g}^{(k)}_t \coloneqq 
\frac{\mathbf{p}(\mathbf{x}^{(k)}_{t+N})-\mathbf{p}(\mathbf{x}^{(k)}_{t})}
{\left\|\mathbf{p}(\mathbf{x}^{(k)}_{t+N})-\mathbf{p}(\mathbf{x}^{(k)}_{t})\right\|+\epsilon}
\end{equation}
denote its unit terminal displacement direction, where $\epsilon>0$ is a small constant for numerical stability.
The cluster-average direction is computed as
\begin{equation}
\begin{aligned}
\mathbf{v}_{t,c} &\coloneqq \frac{1}{|\mathcal{C}_{t,c}|}\sum_{k\in\mathcal{C}_{t,c}} \mathbf{g}^{(k)}_t,\\
\hat{\mathbf{v}}_{t,c} &\coloneqq \frac{\mathbf{v}_{t,c}}{\|\mathbf{v}_{t,c}\|+\epsilon}.
\end{aligned}
\end{equation}

We then estimate the obstacle motion direction from the most recent $H$ position measurements $\{\mathbf{o}_{t-h}\}_{h=0}^{H-1}$, where $\mathbf{o}_{t-h}\in\mathbb{R}^{p}$ denotes the obstacle position in the same task space as $\mathbf{p}(\cdot)$:
\begin{equation}
\begin{aligned}
\mathbf{v}_{\mathrm{obs},t} &\coloneqq \frac{1}{H-1}\sum_{h=0}^{H-2}\frac{\mathbf{o}_{t-h}-\mathbf{o}_{t-h-1}}{\Delta t},\\
\hat{\mathbf{v}}_{\mathrm{obs},t} &\coloneqq \frac{\mathbf{v}_{\mathrm{obs},t}}{\|\mathbf{v}_{\mathrm{obs},t}\|+\epsilon}.
\end{aligned}
\end{equation}

Finally, we select the cluster whose average direction is most opposite to the obstacle motion:
\begin{equation}
c_t^{\star} \coloneqq \arg\min_{c\in\{1,\dots,M_t\}} \ \hat{\mathbf{v}}_{t,c}^{\top}\hat{\mathbf{v}}_{\mathrm{obs},t}\,
\end{equation}
as illustrated by the blue trajectories in Fig.~\ref{DynamicObs}.
Since both vectors are normalized, the dot product is positive when the cluster moves in a similar direction to the obstacle and negative when it moves in the opposite direction. Minimizing it therefore favors rollouts that move against the obstacle flow, which empirically increases clearance and reduces the risk of closing-speed collisions.

\subsubsection{Within-cluster MPPI update}
Finally, CE-MPPI performs the MPPI update using only the rollouts within the selected cluster $\mathcal{C}_{t,c_t^{\star}}$, as illustrated by the orange trajectories in Figs.~\ref{StaticObs} and~\ref{DynamicObs}. Compared to the standard MPPI weights in~\eqref{eq:mppi-w}, we compute weights normalized over the selected cluster:
\begin{equation}
\tilde{w}^{(k)}_t \coloneqq 
\frac{\exp\!\left(-\frac{1}{\lambda}S^{(k)}_t\right)}
{\sum_{r\in\mathcal{C}_{t,c_t^{\star}}}\exp\!\left(-\frac{1}{\lambda}S^{(r)}_t\right)},
\qquad k\in\mathcal{C}_{t,c_t^{\star}},
\end{equation}
and update the nominal control sequence as
\begin{equation}
\mathbf{u}_{t+j} \coloneqq \mathbf{u}_{t+j} + \sum_{k\in\mathcal{C}_{t,c_t^{\star}}} \tilde{w}^{(k)}_t\,\boldsymbol{\epsilon}^{(k)}_{t+j},
\qquad j=0,\dots,N-1.
\end{equation}
As in standard MPPI, only the first control $\mathbf{u}_{t}$ is applied, and the process repeats in a receding-horizon manner.

\section{Experiments}
This section evaluates the effectiveness of the proposed CE-MPPI through both simulation and real-world experiments. In simulation, we construct two 2-D planar scenarios that are prone to failure modes of standard MPPI: a static-obstacle scenario designed to induce averaging-induced failure, and a dynamic-obstacle scenario in which the obstacle motion is aligned with the robot’s nominal direction of travel. These settings assess CE-MPPI’s ability to avoid collisions while maintaining planning efficiency. In real-world experiments, we deploy CE-MPPI on a 6-DoF manipulator in obstacle avoidance and goal reaching tasks, demonstrating that the proposed method remains flexible and efficient even for high-DoF systems.

\subsection{Simulation Setup}
\label{subsec:sim-setup}
% Our simulation is built on the open-source CSC-MPPI implementation \cite{park2025csc}.
We implement CE-MPPI and evaluate it in two planar scenarios: (i) a static-obstacle case designed to induce averaging-induced failure in standard MPPI, and (ii) a dynamic-obstacle case where the obstacle motion is aligned with the robot's nominal direction of travel.

\subsubsection{Hardware and implementation}
All simulation experiments are conducted on a PC equipped with an Intel Core i7-9750H CPU, an NVIDIA GeForce GTX 1650 GPU, and 32\,GB RAM. We implement MPPI in JAX and accelerate rollout simulation and cost evaluation using just-in-time (JIT) compilation.

\subsubsection{Robot model}
The robot is a differential-drive platform with state $\mathbf{x}_t=[x_t,\,y_t,\,\theta_t]^\top$, where $(x_t,y_t)$ denotes the planar position and $\theta_t$ denotes the heading angle. The control input is $\mathbf{u}_t=[u_{v,t},\,u_{\omega,t}]^\top$, where $u_{v,t}$ and $u_{\omega,t}$ are the linear and angular velocities, respectively. Using a discrete-time kinematic model with time step $dt$, the dynamics are
\begin{equation}
\begin{aligned}
x_{t+1} &= x_t + u_{v,t}\cos\theta_t\, dt,\\
y_{t+1} &= y_t + u_{v,t}\sin\theta_t\, dt,\\
\theta_{t+1} &= \theta_t + u_{\omega,t}\, dt.
\end{aligned}
\end{equation}

\subsubsection{Cost function}
We use a quadratic tracking cost with a terminal cost. The stage cost is defined as
\begin{equation}
\ell(\mathbf{x}_t) = (\mathbf{x}_t-\mathbf{x}_f)^\top Q(\mathbf{x}_t-\mathbf{x}_f),
\end{equation}
and the terminal cost is
\begin{equation}
\phi(\mathbf{x}_{t+N}) = (\mathbf{x}_{t+N}-\mathbf{x}_f)^\top H(\mathbf{x}_{t+N}-\mathbf{x}_f),
\end{equation}
where $\mathbf{x}_f$ is the goal state. Similar to  CSC-MPPI~\cite{park2025csc}, we set $Q=\mathrm{diag}(10,10,0)$ and $H=\mathrm{diag}(50,50,50)$.
The other key simulation parameters are summarized in Table~\ref{tab:sim-params}.

\begin{table}[t]
\centering
\caption{Simulation parameters.}
\label{tab:sim-params}
\setlength{\tabcolsep}{6pt}
\renewcommand{\arraystretch}{1.05}
\begin{tabular}{l c}
\hline
\textbf{Parameter} & \textbf{Value} \\
\hline
Rollouts $K$ & 300 \\
Horizon $N$ & 30 \\
Time step $dt$ & 0.03 s \\
Linear velocity bound $|u_{v,t}|$ & $\le 0.8$ m/s \\
Angular velocity bound $|u_{\omega,t}|$ & $\le 7$ rad/s \\
Temperature $\lambda$ & Scenario 1: 0.7,\;\; Scenario 2: 0.01 \\
\hline
\end{tabular}
\vspace{-0.8em}
\end{table}

\subsection{Simulation Results}
\label{subsec:sim-results}
In this section, we conduct comparative simulation experiments in two scenarios. For each scenario, we evaluate three methods: standard MPPI, CSC-MPPI, and the proposed CE-MPPI. All reported quantitative results are averaged over three runs.

\subsubsection{Scenario 1 (static obstacle, averaging-induced failure)}
\begin{figure*}[t]
    \centering
    \includegraphics[width=0.85\textwidth]{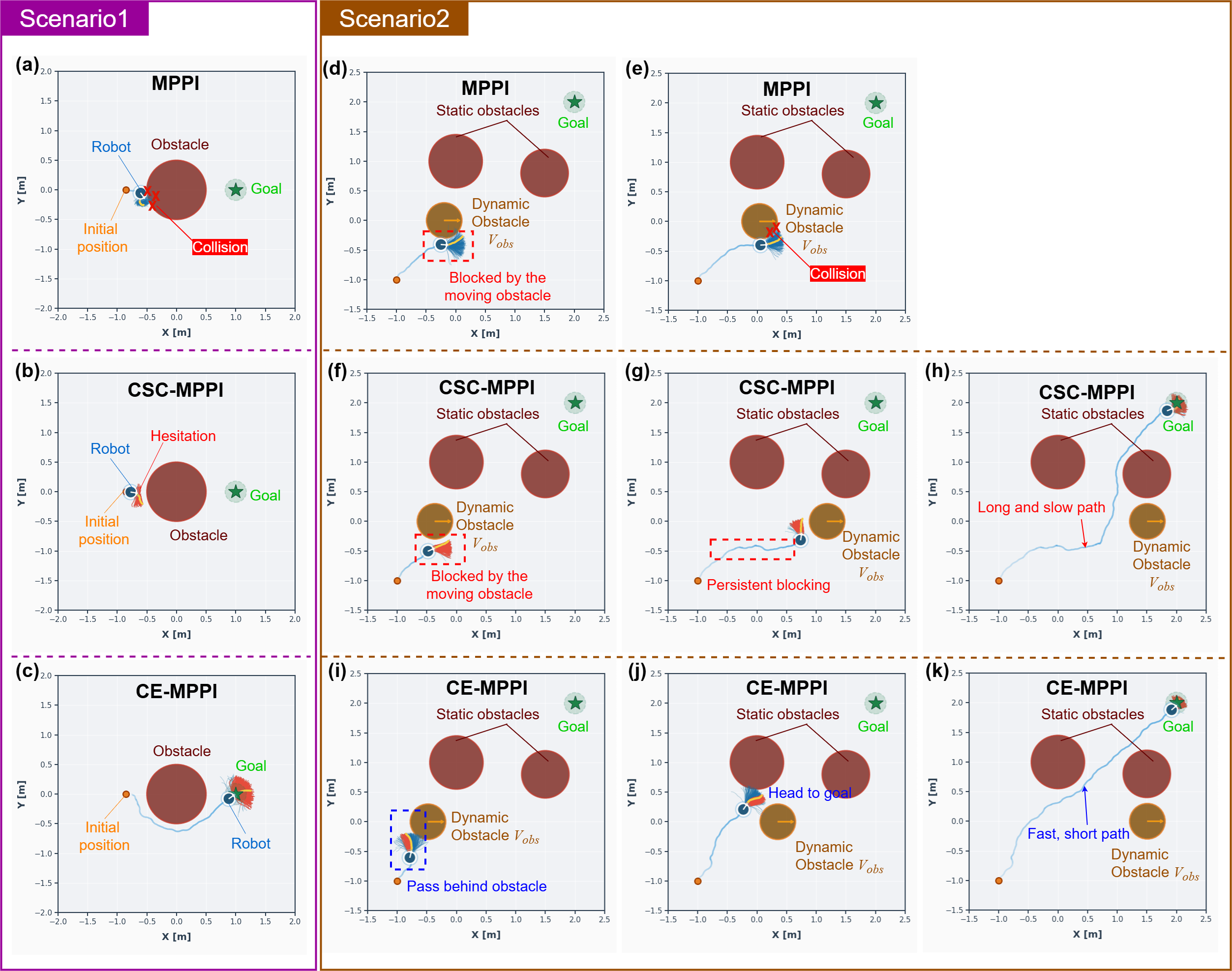}
    \caption{Simulation results in two scenarios. Scenario~1 (static obstacle) is shown in (a), (b), and (c) for MPPI, CSC-MPPI, and CE-MPPI, respectively. Scenario~2 (one dynamic obstacle with two static obstacles) is shown in (d)--(k): MPPI in (d) and (e), CSC-MPPI in (f), (g), and (h), and CE-MPPI in (i), (j), and (k).}
    \label{fig:sim-results}
\end{figure*}

Scenario~1 contains a static obstacle configuration designed to test the ability to avoid averaging-induced failure. The qualitative trajectories are shown in Fig.~\ref{fig:sim-results}(a), (b), and (c), and the quantitative results are summarized in Table~\ref{tab:sim-results}. As shown in Fig.~\ref{fig:sim-results}(a), MPPI can collide with the obstacle in this scenario. This is partly because the robot may build up a relatively high speed while it is still some distance away from the obstacle. When the robot approaches the obstacle, the importance-weighted update can average incompatible avoidance rollouts, producing a control update that still points toward the obstacle. In such cases, the robot may not fully stop in time, resulting in collision.

CSC-MPPI avoids collision in most runs and mitigates averaging-induced failure, but it can exhibit prolonged hesitation near the obstacle, as shown in Fig.~\ref{fig:sim-results}(b). This behavior is related to its clustering criterion based on averaged control statistics, namely the mean linear velocity $v$ and mean angular velocity $\omega$. Even when CSC-MPPI separates collision-free rollouts into two clusters that pass on different sides of the obstacle, the mean $v$ across clusters can remain similar, which reduces cluster separability and can delay decisive avoidance.

In contrast, CE-MPPI clusters feasible rollouts using a geometric feature derived from the direction vector from the collision-derived reference point to each rollout terminal position. This feature more clearly distinguishes clusters with different avoidance directions, rather than clustering based on velocity magnitude. As a result, even when feasible rollouts form two clusters on opposite sides of the obstacle, CE-MPPI can efficiently select one cluster and proceed toward the goal, as shown in Fig.~\ref{fig:sim-results}(c). Quantitatively, Table~\ref{tab:sim-results} shows that CE-MPPI reaches the goal faster than CSC-MPPI and with a slightly shorter path length in Scenario~1. Since MPPI is not collision-free in this scenario, no statistics are available there.

\begin{table}[t]
\centering
\caption{Simulation results in two scenarios.}
\label{tab:sim-results}
\setlength{\tabcolsep}{5pt}
\renewcommand{\arraystretch}{1.05}
\begin{tabular}{c l c c c}
\hline
\textbf{Scenario} & \textbf{Method} & \textbf{Collision-free} & \textbf{Time (s)} & \textbf{Path (m)} \\
\hline
\multirow{3}{*}{1} 
& MPPI     & No & -- & -- \\
& CSC-MPPI & Yes & 4.23 & 2.27 \\
&  \textbf{CE-MPPI}  & Yes & \textbf{4.02} &  \textbf{2.22} \\
\hline
\multirow{3}{*}{2}
& MPPI     & No & -- & -- \\
& CSC-MPPI & Yes & 9.66 & 4.73 \\
&  \textbf{CE-MPPI}  & Yes &  \textbf{7.98} &  \textbf{4.30} \\
\hline
\end{tabular}
\end{table}

\subsubsection{Scenario 2 (dynamic obstacle with static obstacles)}
Scenario~2 includes one dynamic obstacle and two static obstacles, as shown in Fig.~\ref{fig:sim-results}(d)--(k) and Table~\ref{tab:sim-results}. The dynamic obstacle (yellow) moves along the positive $x$ direction at approximately $0.43$~m/s and interferes with the robot as it travels toward the goal. Both MPPI and CSC-MPPI are affected by the moving obstacle. For MPPI, when the robot approaches the obstacle, it tends to steer to the right to avoid collision while still attempting to make progress toward the goal. However, the moving obstacle continues to block the robot, and in some cases MPPI collides with the obstacle, as shown in Fig.~\ref{fig:sim-results}(d) and (e). CSC-MPPI avoids collision but still suffers from persistent blocking. As illustrated by the long segment of robot motion roughly along the $x$ axis in Fig.~\ref{fig:sim-results}(f), (g), and (h), the robot continues to move while waiting for the moving obstacle to clear before committing to a detour that reduces the cost. As a result, CSC-MPPI remains blocked for an extended period and travels an unnecessarily long path, which is reflected by the $9.66$~s travel time and $4.73$~m path length in Table~\ref{tab:sim-results}.

In contrast, CE-MPPI explicitly accounts for the obstacle motion direction during cluster selection. Once a moving obstacle is detected, CE-MPPI selects the rollout cluster whose average motion is opposite to the obstacle motion direction, enabling an early bypass behind the obstacle, as shown in Fig.~\ref{fig:sim-results}(i). Although this choice may not minimize the instantaneous cost at the current step, incorporating obstacle direction provides a simple predictive bias that reduces persistent blocking and improves overall planning efficiency. This is corroborated by the shorter travel time ($7.98$~s) and shorter path length ($4.30$~m) achieved by CE-MPPI in Table~\ref{tab:sim-results}.

\subsection{Real-World Experimental Setup}
\label{subsec:real-setup}

\begin{figure}[htbp]
    \centering
    \includegraphics[width=1.0\linewidth]{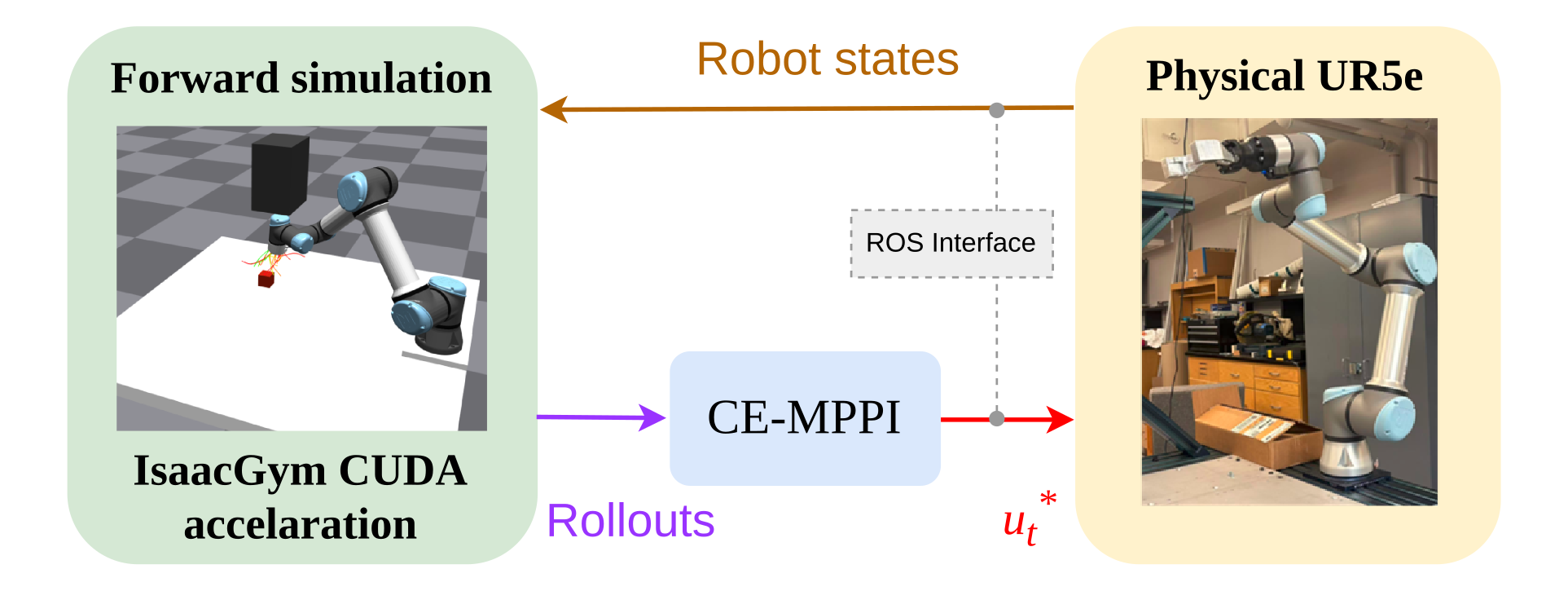}
    \caption{Experimental setup.}
    \label{experiment_setup}
\end{figure}

To validate CE-MPPI in real-world settings and examine its effectiveness on high-DoF robots, we conduct experiments on a 6-DoF UR5e manipulator. The experimental platform is shown in Fig.~\ref{experiment_setup}. We use the same PC as in the simulation as the onboard computing platform. Different from the 2-D simulation, forward simulation for MPPI rollouts is performed in Isaac Gym, leveraging CUDA-based parallel computation to improve runtime efficiency.

\subsubsection{State and control interface.}
We read the UR5e joint states via ROS, using a 12-dimensional state vector
$\mathbf{x}_t=[\mathbf{q}_t^\top,\dot{\mathbf{q}}_t^\top]^\top\in\mathbb{R}^{12}$,
where $\mathbf{q}_t\in\mathbb{R}^{6}$ and $\dot{\mathbf{q}}_t\in\mathbb{R}^{6}$ denote joint positions and velocities. CE-MPPI outputs a joint-velocity command $\mathbf{u}_t\in\mathbb{R}^{6}$, which is sent back to the UR5e through ROS for receding-horizon execution. Due to GPU memory constraints, we set the control rate to 10\,Hz ($dt=0.1$\,s). The other key parameters of the real-world experiment are summarized in Table~\ref{tab:real-params}.

\begin{table}[t]
\centering
\caption{Real-world experiment parameters}
\label{tab:real-params}
\setlength{\tabcolsep}{5pt}
\renewcommand{\arraystretch}{1.05}
\begin{tabular}{l c}
\hline
\textbf{Parameter} & \textbf{Value} \\
\hline
Rollouts $K$ & 50 \\
Horizon $N$ & 12 \\
Control step $dt$ & 0.1 s (10 Hz) \\
Joint velocity bound $|\dot{q}_i|$ & $\le 0.1$ rad/s \\
Joint position bound $|q_i|$ & $\le 2\pi$ rad \\
Temperature $\lambda$ & 0.05 \\
DBSCAN $\varepsilon$ (task space) & 0.3 m \\
\hline
\end{tabular}
\vspace{-0.6em}
\end{table}

\subsubsection{Cost function}
We use a goal-reaching objective with collision penalties. Let $\mathbf{p}_{ee}(\mathbf{q}_t)\in\mathbb{R}^3$ and $R_{ee}(\mathbf{q}_t)\in SO(3)$ denote the end-effector position and orientation from forward kinematics, and let $(\mathbf{p}_g,R_g)$ be the goal pose. The stage cost is defined as
\begin{equation}
\begin{aligned}
\ell(\mathbf{x}_t,\mathbf{u}_t)
&= w_p\left\|\mathbf{p}_{ee}(\mathbf{q}_t)-\mathbf{p}_g\right\|_2
+ w_R\, d_R\!\big(R_{ee}(\mathbf{q}_t),R_g\big) \\
&\quad + w_c\, \mathbb{I}_{\mathrm{col}}(\mathbf{q}_t),
\end{aligned}
\end{equation}
where $d_R(\cdot,\cdot)$ measures orientation mismatch (e.g., quaternion angle error), and $\mathbb{I}_{\mathrm{col}}(\mathbf{q}_t)\in\{0,1\}$ indicates collision or contact with the table, shelf and obstacle. For the UR5e robot and our task space, we set $(w_p,w_R,w_c)=(10,3,1000)$.

\subsection{Real-World Experimental Results}
\label{subsec:real-results}

\begin{figure}[htbp]
    \centering
    \includegraphics[width=0.85\linewidth]{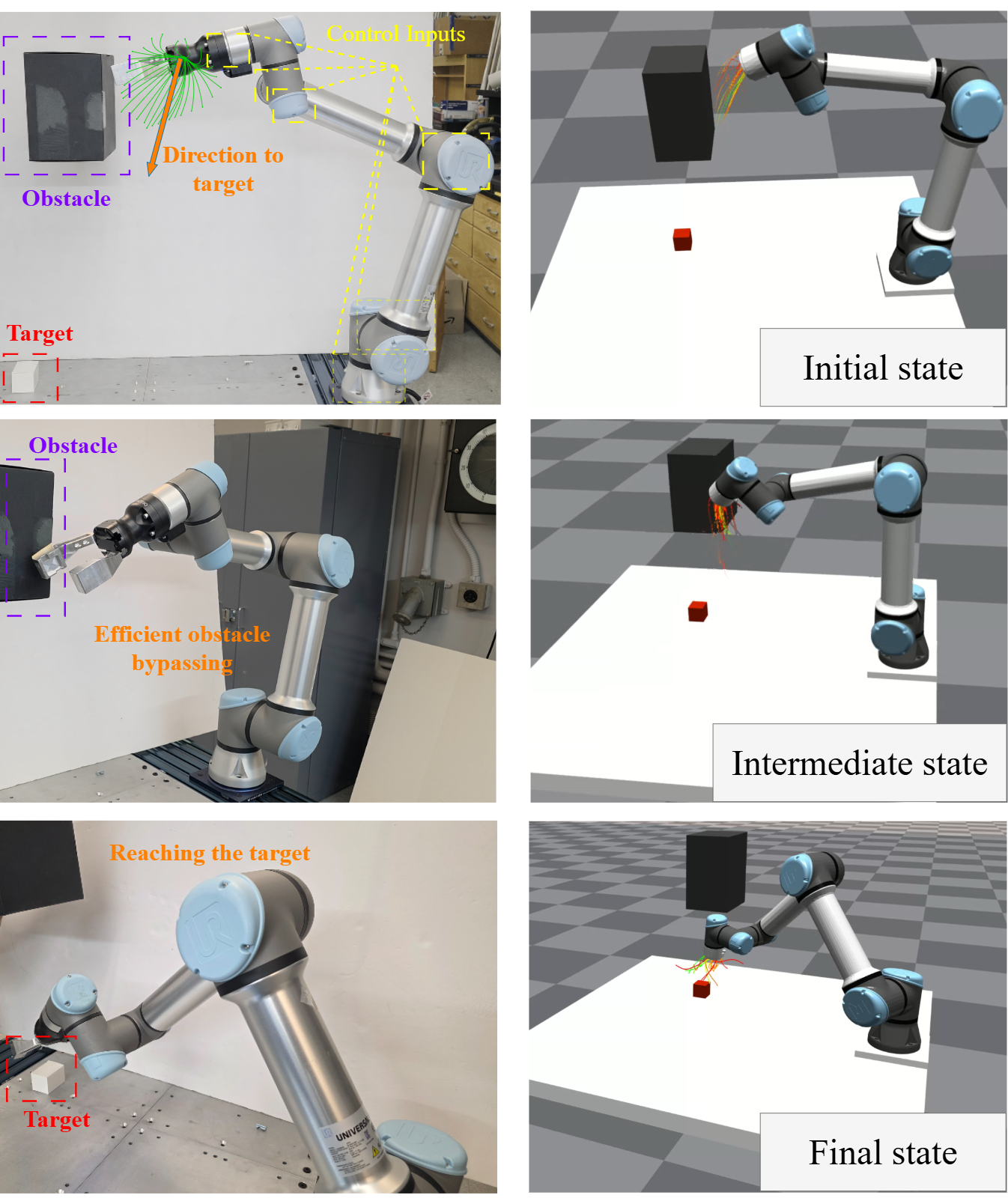}
    \caption{Real-world UR5e tabletop experiment. A black box serves as the obstacle and the cube on the table is the end-effector goal target.}
    \label{experiments}
\end{figure}

We conduct real-world experiments on the UR5e platform shown in Fig.~\ref{experiments}. The scene is set on a tabletop with a black box placed in front of the robot as an obstacle. A cube on the table serves as the goal target for the end-effector to reach.

\subsubsection{Baseline feasibility}
CSC-MPPI requires iteratively modifying colliding rollouts via gradient-based updates to move them out of collision regions. On a high-DoF manipulator, where sampling perturbs all joints simultaneously, this refinement step introduces substantial computational overhead. In our setup, a single planning cycle exceeds the 10\,Hz control budget, preventing CSC-MPPI from producing control commands in time for real-time execution. Therefore, we focus on comparing standard MPPI and CE-MPPI on the real robot.

\subsubsection{MPPI vs. CE-MPPI}
For both MPPI and CE-MPPI, the collision penalty is set sufficiently large so that the robot remains collision-free in practice. However, standard MPPI often exhibits prolonged hesitation near the obstacle, which significantly increases the time to reach the goal. In contrast, CE-MPPI performs the weighted update using only rollouts from the selected cluster, avoiding averaging across incompatible avoidance rollouts and enabling faster escape from the near-obstacle stagnation region.
Table~\ref{tab:real-results} summarizes the results. MPPI takes 179.20~s to reach the target, nearly twice as long as CE-MPPI (93.11~s), and yields a longer end-effector path (0.83~m vs. 0.73~m). These results indicate that CE-MPPI improves escape efficiency and goal-reaching performance in the presence of nearby obstacles on a high-DoF manipulator.

\begin{table}[t]
\centering
\caption{Real-world results on the UR5e tabletop setup.}
\label{tab:real-results}
\setlength{\tabcolsep}{6pt}
\renewcommand{\arraystretch}{1.05}
\begin{tabular}{l c c c}
\toprule
\textbf{Method} & \textbf{Collision-free} & \textbf{Time (s)} & \textbf{EE Path (m)} \\
\midrule
MPPI    & Yes & 179.20 & 0.83 \\
\textbf{CE-MPPI} & Yes & \textbf{93.11}  & \textbf{0.73} \\
\bottomrule
\end{tabular}
\vspace{-0.6em}
\end{table}

\section{Conclusions}
We propose CE-MPPI to architecturally lift MPPI by pruning colliding rollouts and optimally clustering feasible rollouts using a compact geometric feature. CE-MPPI then selects a single cluster for a within-cluster MPPI update, using a minimum-cost criterion for static obstacles and an opposite-to-obstacle-motion criterion for dynamic obstacles. This design mitigates averaging-induced failure and reduces persistent blocking by proactively choosing a bypass direction in dynamic scenes. Simulation on a differential-drive robot and real-world experiments on a 6-DoF UR5e manipulator demonstrate the efficiency and effectiveness of CE-MPPI for real-time motion planning.

\addtolength{\textheight}{-12cm}   % This command serves to balance the column lengths

%%%%%%%%%%%%%%%%%%%%%%%%%%%%%%%%%%%%%%%%%%%%%%%%%%%%%%%%%%%%%%%%%%%%%%%%%%%%%%%%

%%%%%%%%%%%%%%%%%%%%%%%%%%%%%%%%%%%%%%%%%%%%%%%%%%%%%%%%%%%%%%%%%%%%%%%%%%%%%%%%



% \section*{ACKNOWLEDGMENT}
% Support for this research was provided in part by an Amazon Science-Hub gift award. The views and conclusions contained in this document are those of the authors and should not be interpreted as representing the official policies, either expressed or implied, of the sponsoring organization or the University of Washington.

%\begin{thebibliography}{99}

\bibliographystyle{IEEEtran}
\bibliography{IROS2026Ref_Liuzd}

%\end{thebibliography}

\end{document}